# Mathematical Principles and Experimental Discoveries of the Emergence of Symbolic Patterns in Artificial Neural Networks

Quanshi Zhang*, Qihan Ren*, Siyu Lou
School of Computer Science, Shanghai Jiao Tong University, Shanghai, China

Correspondence to: Quanshi Zhang (zqs1022@sjtu.edu.cn)

**Abstract**

Artificial Neural networks (ANNs) are often treated as black-box models, making explainability a central challenge in deep learning. Many engineering methods have been proposed to approximately explain the ANN from various perspectives, such as feature attribution and visualization. However, it remains a long-standing open question whether the complex inference logic of an ANN can be explained exhaustively and concisely as sparse symbolic patterns. This raises a deeper inquiry: does the emergence of symbolic patterns reflect a natural law rather than chance? Here, we show that across a broad class of ANNs trained on diverse tasks, their inference logic can indeed be reformulated as sparse symbolic interactions. We further prove that two common mathematical criteria, which are implicitly required across tasks, lead to the emergence of such sparse symbolic interactions. Empirical evidence confirms that the two criteria hold for the majority of input samples in diverse models. Furthermore, the faithfulness of these interactions is also demonstrated by their strong sample-to-sample and model-to-model transferability, as well as their ability to explain the overall generalization power of ANNs. Our theoretical analysis and extensive experiments provide a solid foundation for symbolic explanations of ANNs, and offer novel insights into the ANN's generalization power. Our findings also highlight the potential of *communicative learning*, a paradigm in which the inference logic of an ANN can be directly inspected and tuned at the level of symbolic patterns, thus complementing traditional end-to-end learning paradigm. Finally, the observed emergence of symbolic patterns in ANNs suggests that similar symbolic representations may also emerge in other types of black-box systems under certain conditions, because our proof does not depend on any specific ANN architecture.

Emergence describes the spontaneous formation of novel properties in complex systems that cannot be predicted from their individual components. This phenomenon, extensively studied across disciplines[1–4], manifests in diverse forms such as physical phase transitions[5] and biological self-organization in insect societies[6], where simple patterns or rules arise from intricate networks or linking structures. A particularly intriguing yet unresolved question is how symbolic patterns[7,8] emerge in neural networks. While it remains intractable to develop a fully analytical model of biological neural systems[9], artificial neural networks (ANNs) offer a computationally feasible alternative for studying this phenomenon. Notably, some studies[10] suggest that Large Language

Models (LLMs) may partially encode symbolic logic, although the view remains debated.

Therefore, in this study, we re-examine the emergence of symbolic patterns in ANNs, addressing three key questions:

1. Can symbolic patterns spontaneously emerge in an ANN?
2. What mathematical criteria can lead to the emergence of symbolic patterns?
3. Do the emergent symbolic patterns represent transferable knowledge?

Given an input sample $x$ comprising $n$ variables\footnote{For instance, each input variable may represent an image patch or the embedding vector of a word/token in a text sequence.}, we consider an ANN $v$ that produces a scalar output $v(x)$. While current theoretical frameworks primarily address networks with scalar outputs, this setting is practical and can be readily extended to multi-dimensional outputs.\footnote{For example, in multi-class classification tasks, the scalar output $v(x)$ can be defined as either the pre-softmax logit or the log odds of the target class.}

Our basic idea is to use the following AND-OR logical model $h$ based on symbolic interactions to explain the inference logic of the ANN:

$$h(x') := \sum_{S\in\Omega_{\text{and}}} I_S^{\text{and}} \cdot \mathbb{1}_{\text{and}}(S|x') + \sum_{S\in\Omega_{\text{or}}} I_S^{\text{or}} \cdot \mathbb{1}_{\text{or}}(S|x'),$$

where $x'$ denotes a masked state of the original input $x$, and any subset of input variables may be masked.\footnote{For instance, on text inputs, we may replace the tokens to be masked with a padding token or a certain embedding vector.} The binary trigger function $\mathbb{1}_{\text{and}}(S|x')$ indicates an AND interaction over the subset $S$. Specifically, this function returns 1 if and only if all variables in $S$ remain unmasked in $x'$; otherwise, it returns 0. Similarly, $\mathbb{1}_{\text{or}}(S|x')$ represents an OR interaction, returning 1 if at least one variable in $S$ is unmasked in $x'$. Accordingly, scalar weights $I_S^{\text{and}}$ and $I_S^{\text{or}}$ represent the interaction effects. The *order* of an interaction on set $S$ is defined as the cardinality $|S|$.

In this section, we demonstrate the faithfulness of the explanation based on the logical model $h$ from two perspectives. (1) **Fidelity**: We prove the *universal matching* property of interactions in the logical model $h$, meaning that the logical model can accurately explain the network outputs of the ANN on an exponential number of samples. (2) **Conciseness**: We show the *sparsity* property of interactions, meaning that the logical model contains only a small number of AND-OR interactions.

**Universal matching property**: A logical model $h$ is constructed to explain the inference logic of a specific input sample $x$ with $n$ variables. We prove that by configuring a specific set of interaction effects $I_S^{\text{and}}$ and $I_S^{\text{or}}$ (please see **Methods** for computational details), the output of the logical model $h(x')$ can precisely approximate the scalar output $v(x')$ of the ANN across all possible masked states $x'$ of the input sample $x$. Specifically, as **Error! Reference source not found.** shows, let us consider a masked state $x'$of the input sample $x$ from all its $2^n$ possible masked configurations. Each masked state $x'$triggers a specific set of AND and OR interactions, which contribute their numerical effects, $I_S^{\text{and}}$ and $I_S^{\text{or}}$, additively to the output of the logical model $h(x')$. The output $h(x')$ closely matches the ANN's output $v(x')$, which ensures the fidelity of the explanation.

**Sparsity Property:** We observe that the logical model $h$ typically encodes only a sparse subset of interactions, and a concise logical model with these sparse interactions is powerful enough to approximate all network outputs on all $2^n$ masked states. In fact, we identify two mathematical criteria that usually lead to the sparsity (conciseness) of interactions: monotonicity and smoothness of the ANN's average $k$-order output. The two criteria are commonly satisfied by ANNs and are introduced later.

The two properties above demonstrate that a concise logical model can universally mimic the outputs of an ANN across a wide range of augmented samples. *This suggests that we can regard AND and OR interactions as primitive inference patterns encoded by the ANN.*

Do the primitive inference patterns of ANNs reside in the activation of specific neurons, or do they arise from the complex interactions within the network? Our finding differ from mechanistic interpretability studies[11–13], which investigate the roles of individual neurons or circuits in an ANN. In contrast to this line of work, the interactions we study are encoded by the entire ANN, rather than being localized to specific neurons. Furthermore, subsequent sections show that such interactions are transferable between different samples and models, whereas individual neurons do not exhibit similar transferability. This suggests that interactions may represent a more fundamental aspect of neural network representations, rather than being tied to specific individual neurons.

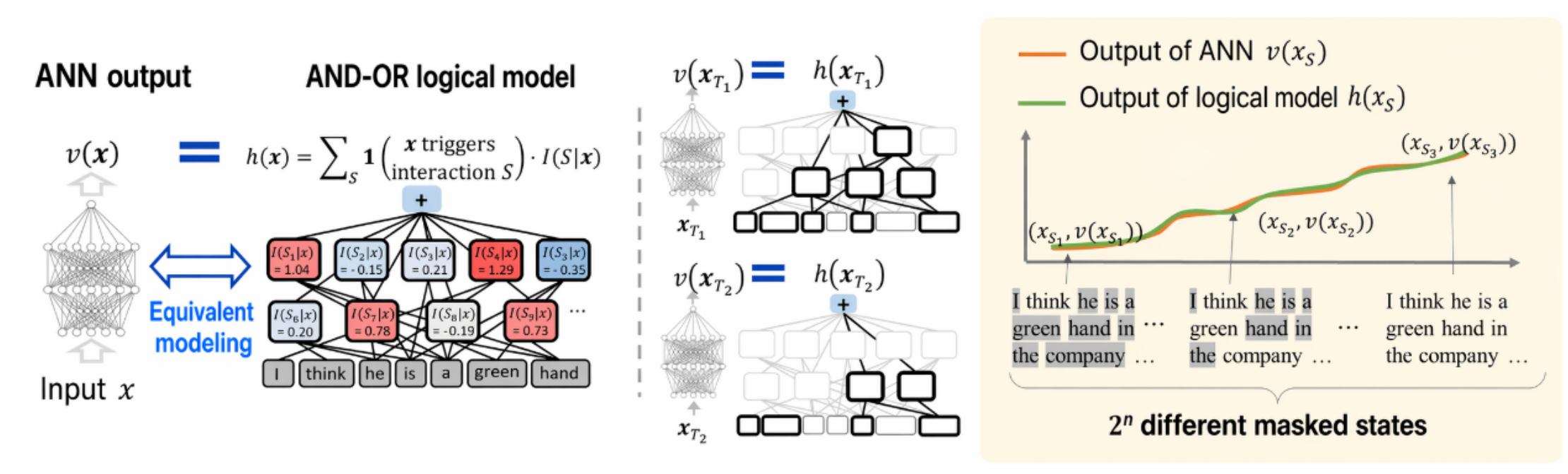

*Fig. 1 | Overview of the AND–OR logical model for explaining ANN inference.* ***a,*** *The inference logic of an ANN on a specific input sample is represented by a sparse logical model composed of AND and OR interactions. Each interaction is triggered by the presence or absence of a particular subset of input variables and contributes its corresponding interaction effect to the model output.* ***b,*** *Universal matching property. For each masked state of the input, the triggered AND and OR interactions are aggregated to produce the logical model output $h(x_T)$, which closely matches the ANN output $v(x_T)$. Thus, a small set of interactions can faithfully explain the ANN outputs over all $2^n$ possible masked states.*

## Discovery and Mathematical Principles of Interaction Sparsity

**Emergence of sparse interactions across diverse ANNs and tasks.** The sparsity of AND-OR interactions is empirically observed across a broad range of ANNs trained on various tasks. As shown in Fig. 1 , we analyze the sparsity of interactions extracted from Convolutional Neural Networks (CNNs) for image recognition, LLMs for text generation, and the PointNet[14] for 3D point cloud classification. To facilitate the computation of interactions, we select 10 image patches from the input image as input variables for CNNs, select 10 words or phrases from the input prompt for LLMs, and select 9 point cloud regions from the input 3D point cloud for PointNet.

We find that although there are $2^{11}$ total interaction effects ($I_S^{\text{and}}$ and $I_S^{\text{or}}$) for different subsets $S \subseteq N$ on CNNs and LLMs ($2^{10}$ on the PointNet), only a small number of these interactions have salient absolute effects. The number of salient interactions (averaged over samples) ranges from a minimum of 10.2 to a maximum of 79.1 across different models we tested, with all other interactions having negligible effects. This is termed *the sparsity of interactions*.

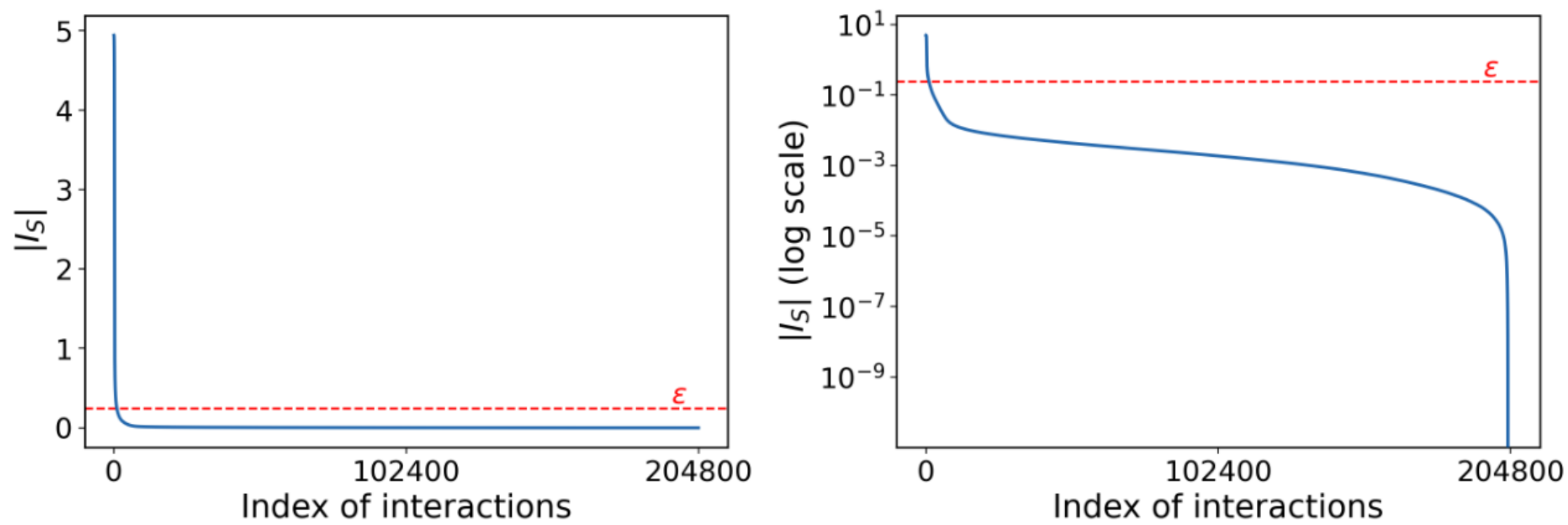

***Fig. 1 | Sparsity of interaction effects.*** *We examine AND-OR interactions on CNNs for image recognition (VGG-16[15], ResNet-18/34/50[16]), LLMs for text generation (Qwen2.5-0.5B/1.5B/3B/7B/14B[17], Deepseek-R1-distilled-1.5B/8B/14B[18], Phi-4[19]), and PointNet for 3D point cloud classification. For each model, we show the interaction effects on all samples, sorted by their absolute values ($|I_S^{and}|$ and $|I_S^{or}|$) in descending order.*

The above finding implies that we can use only a small number of salient interactions to construct the logical model. We find that the constructed logical model can still closely match the network outputs on all $2^n$ masked states. Fig. 2 shows the fidelity of using the output of this logical model to match the ANN's outputs on all $2^n$ masked states.

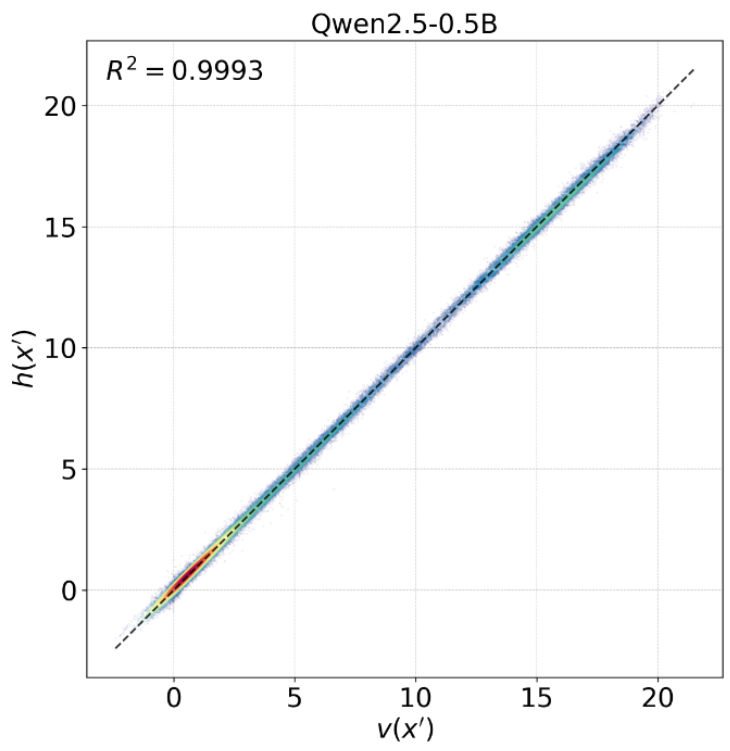

***Fig. 2 | Fidelity of the AND-OR logical model.*** *The scatter plot shows the fidelity of using the logical model's output to match the ANN's output. Each data point in the plot represents a specific masked state on a specific input*

*sample. The x-coordinate denotes the ANN's output* $v(x')$*. The y-coordinate denotes the logical model's output* $h(x')$ *when we only use 100 interactions with top-ranked absolute effects. The coefficient of determination* $R^2$ *is 0.997±0.002 across different models, indicating a good match.*

This sparsity of interactions raises a natural question: is it simply an artifact of model size? One might hypothesize that smaller ANNs usually encode sparser interactions, or conversely, that the suboptimality of feature representations in smaller ANNs leads to numerous redundant or noisy interactions. As shown in Fig. 1, the interaction sparsity remained relatively stable across all model sizes. This result suggests that sparsity is an intrinsic organizing principle of the network's learned logic, rather than a byproduct of its size.

**Mathematical criteria for the emergence of sparse interactions.** Given a specific input sample $x$, we aim to explain the reason for the interaction sparsity, i.e., why the ANN only uses a small number of interactions for inference. To facilitate our analysis, we restrict our attention to ANN $v$ and input sample $x$ that meet the following two criteria.

***Criterion 1 (Monotonicity)*:** Given an input sample $x$ with $n$ input variables, let $\bar{v}^{(k)}(x)$ denote the *average* $k$*-order output* on this sample. $\bar{v}^{(k)}(x)$ is calculated by averaging the network outputs $v(x')$ over all $\binom{n}{k}$ possible states $x'$ where exactly $k$ of the original $n$ input variables are kept unmasked. The monotonicity criterion requires that $\bar{v}^{(k)}(x)$ increases monotonically as the number of unmasked variables $k$ increases. In short, on average, the model's classification confidence should not decrease as more input variables are unmasked.

***Criterion 2 (C-Smoothness)*:** The average $k$-order output $\bar{v}^{(k)}(x)$ changes smoothly with $k$. Formally, the smoothness criterion requires that the second-order finite difference of $\bar{v}^{(k)}(x)$ with respect to the number of unmasked variables $k$, is upper-bounded by a constant $\frac{C}{n}$. This implies that the average marginal gain from unmasking an additional variable does not change uncontrollably.

These two criteria allow us to simplify the analysis of the inference of an ANN $v$ on a specific sample $x$. The analysis now relies solely on the sequence of average scores $[\bar{v}^{(0)}(x), \bar{v}^{(1)}(x), \ldots, \bar{v}^{(n)}(x)]$ of all network outputs when we mask different numbers of input variables. Therefore, our analysis does not need to consider excessive complex factors such as network architecture and parameters. We systematically study these sequences under various settings, including the number of input variables $n$, the smoothness constant $C$, and the constraint on matching error $\epsilon$. The chosen range for $C$ aligns with the empirical values observed in real ANNs, while the range of the error constraint $\epsilon$ is set to be small relative to the ANN's output.

For each triplet $(n, C, \epsilon)$, we uniformly sample a set of $\bar{v}^{(k)}(x)$ sequences. For each given sequence, we can derive the upper bound $U$ on the number of salient interactions. This upper bound constrains the maximum number of interactions that have salient effects (those with absolute values exceeding $\epsilon$), while also satisfying the universal matching property within a small error, i.e., for each masked state $x'$ of the input, $|h(x') - v(x')| \leq \epsilon$.

Specifically, for each specific sequence, the upper bound of salient interactions is given by $U = \frac{M}{\eta\epsilon}$, where $\eta$ (between 0 and 1) reflects the cancellation effect among positive and negative interactions. Although we do not theoretically derive the distribution of $\eta$, empirical results in Fig. 3(a) show that $\eta$ is not extremely small. On LLMs, $\eta$ is greater than 0.1 for over 96.6% of the samples, with a mean of 0.376. On CNNs, $\eta$ is greater than 0.1 for over 86.9% of the samples, with a mean of 0.247. Thus, the quantity $M$ can serve as a practical estimator of interaction sparsity. Fig. 4(b) shows the 95th percentile of $M$ across all sequences $\bar{v}^{(k)}(x)$, under different settings of $(n, C, \epsilon)$. Our results show that across different settings of $C$ and $\epsilon$, the 95th percentile of $M$ is at most 7.8 for $n = 10$, 23.3 for $n = 20$, and 108.5 for $n = 40$, which are relatively small. These results for $M$ imply that the upper bound $U = \frac{M}{\eta\epsilon}$ is significantly less than the number of potential AND-OR interaction candidates, which is $2^{n+1}$. This demonstrates that most ANNs achieve universal matching using a remarkably sparse set of interactions.

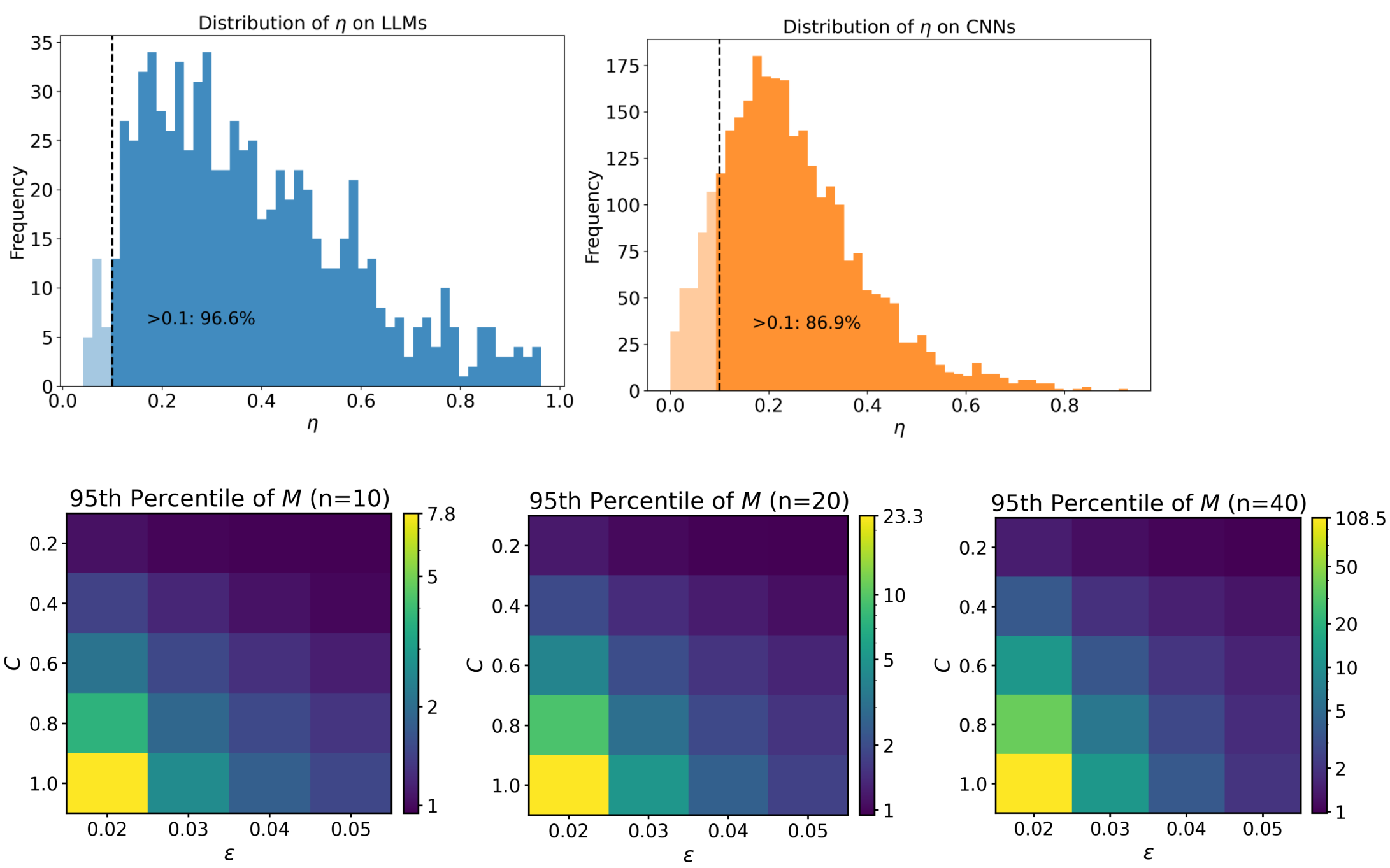

*Fig. 3* | ***Theoretical analysis of interaction sparsity. a,*** *Distributions of the cancellation coefficient $\eta$ on LLMs and CNNs. In most samples, $\eta > 0.1$, indicating limited cancellation between positive and negative interaction effects.* ***b,*** *The 95th percentile of the theoretical upper bound $M$ under different smoothness constants $C$, matching-error tolerances $\epsilon$, and numbers of input variables $n$. The bound remains small relative to the total number of possible interactions.*

**Commonness of the two criteria.** The two criteria reflect properties one would intuitively expect from well-trained ANNs. The first criterion implies that inference confidence should generally increase as more input variables (e.g., words in a sentence or patches in an image) are retained. Anomalies in the DNN outputs can typically be smoothed out, since the model outputs are averaged over a combinatorially large number of masked states that retain $k$ input variables.

The second criterion requires the average $k$-order output $\bar{v}^{(k)}(x)$ to be smooth. When we consider all possible ways of randomly masking $k$ out of $n$ input variables, the average inference confidence is significantly smoothed by the averaging operation. As a result, the average inference confidence of masking $k$ variables does not change abruptly compared to that of masking $k+1$ variables.

## Do Heterogeneous ANNs Share a Set of Universal Optimal Interactions?

**Representation convergence of interactions across different ANNs.** While network architecture and training data are widely seen as the main factors shaping the representations learned by ANNs, interaction patterns provide a more fundamental aspect to analyze an ANN. Given the same input, we discover that different ANNs, which have achieved state-of-the-art performance on the same task, tend to rely on similar interaction patterns for inference.

We conduct comparative experiments to evaluate the sharedness of interactions between two pairs of LLMs: (1) the pair of Qwen2.5-7B and Deepseek-R1-Distill-Llama-8B, and (2) the pair of Phi-4 and Deepseek-R1-Distill-Qwen-14B. As shown in Fig. 4, the two model pairs show a high ratio of shared interactions, with first- and second-order interactions showing the highest sharedness. Since interactions of orders 8 to 10 are extremely sparse, we do not draw strong conclusions from these high-order results. In addition, Fig. 4 illustrates a concrete example on the interactions shared by both the Qwen model and the Deepseek model.

Qwen2.5-7B vs. Deepseek-R1-Distill-Llama-8B

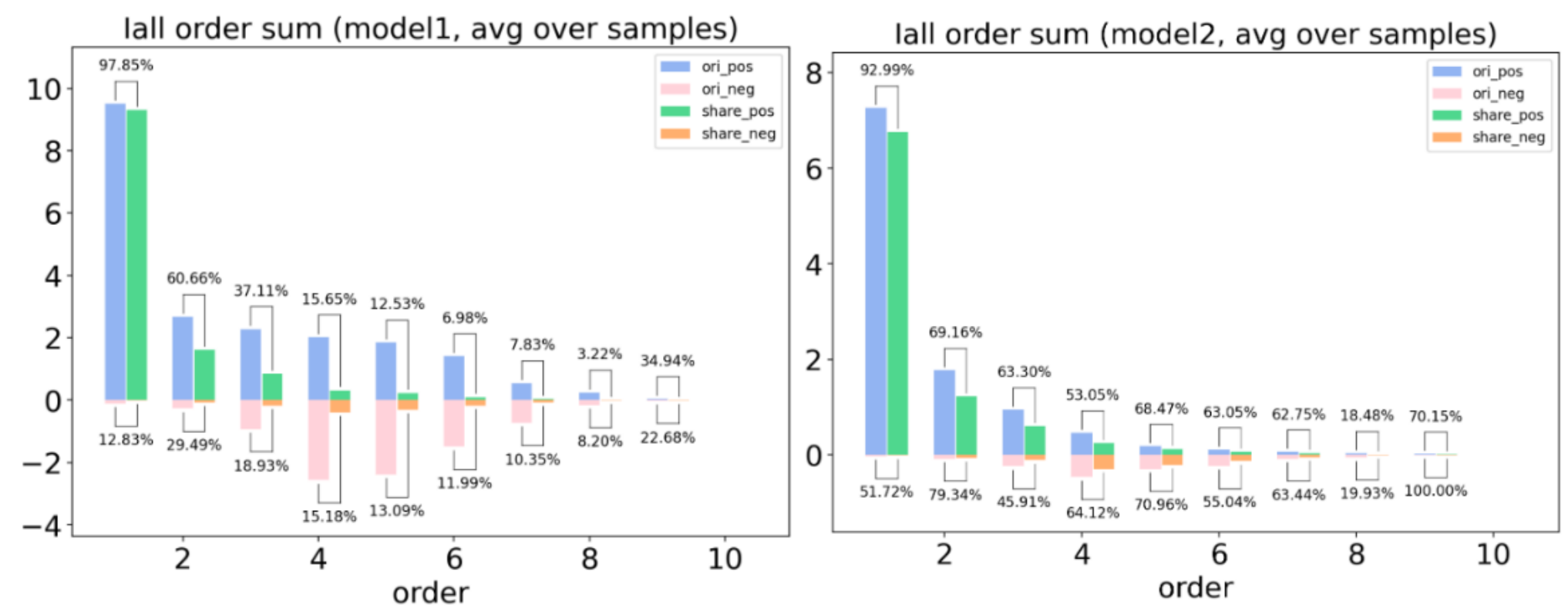

Phi4-14B vs. Deepseek-R1-Distill-Qwen-14B

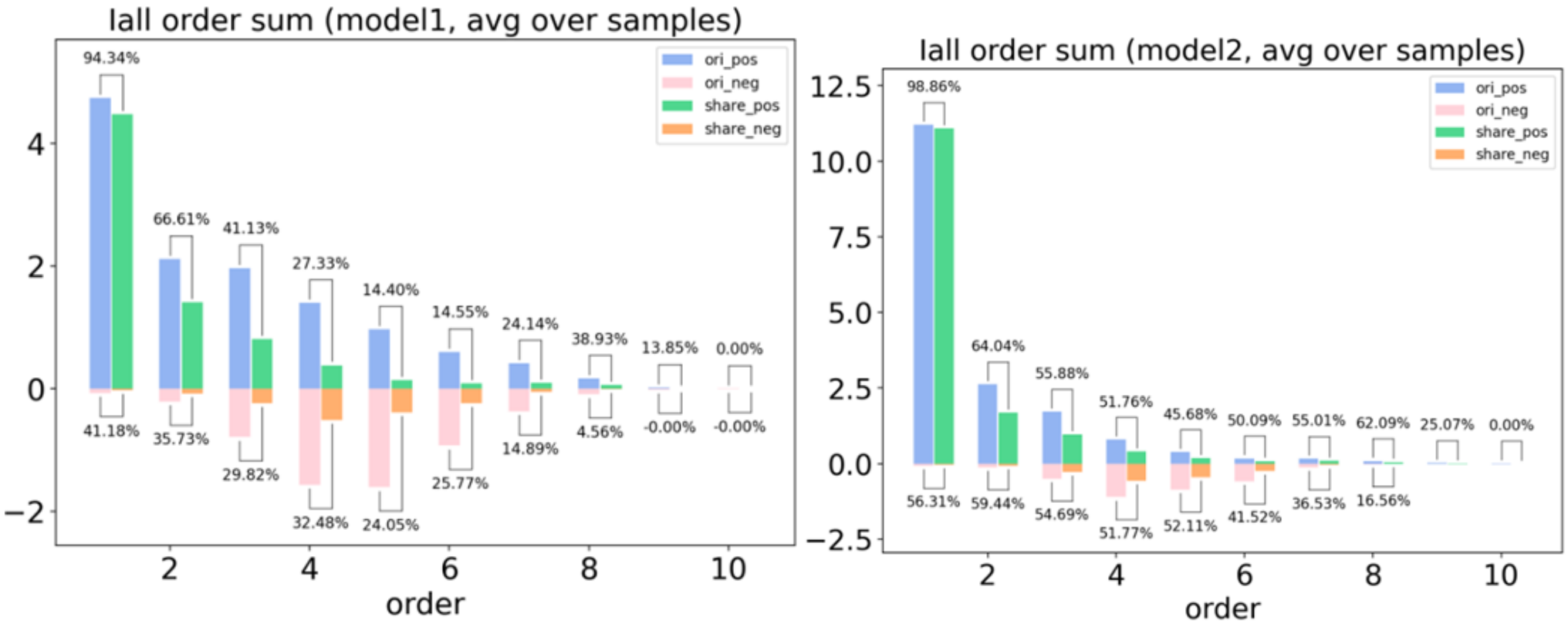

*Fig. 4* | ***Convergence of interaction representations across heterogeneous LLMs.*** *Results compare Qwen2.5-7B with Deepseek-R1-Distill-Llama-8B (top) and Phi-4 with Deepseek-R1-Distill-Qwen-14B (bottom). For each pair, the two panels show the order-wise sums of positive and negative interaction effects in the respective models, decomposed into effects contributed by interactions shared between the two models and effects contributed by model-specific interactions. Percentages indicate the proportion of shared interaction effects at each order. Low-order interactions, especially first- and second-order interactions, exhibit the strongest cross-model sharedness.*

**Transferability of interactions across different samples.** Beyond the representation convergence of interaction patterns across different ANNs, we also observe that many salient interactions extracted from one sample can also be extracted as a salient interaction from another sample in the same category. We refer to this as the *cross-sample transferability* of interactions.

We evaluate such cross-sample transferability of interactions on various ANNs. For ResNet-50 trained to classify the *eyeglasses* attribute on the CelebA dataset[20], we randomly select 50 face images with the *eyeglasses* attribute and compute interactions between nine annotated facial components (Fig. 6). For PointNet trained on ShapeNet dataset[21], we randomly select 100 point clouds in the motorbike category and compute interactions between nine object parts (Fig. 6). In order to evaluate the *cross-sample transferability* of interactions, we construct a dictionary for each ANN, which contains the most frequent salient interactions across different samples in the same category. As shown in Fig. 5, a dictionary containing 100 interactions suffices to cover 83.0% and 68.7% of salient interactions on the CelebA and ShapeNet datasets, respectively. This result underscores the strong transferability of interactions across samples.

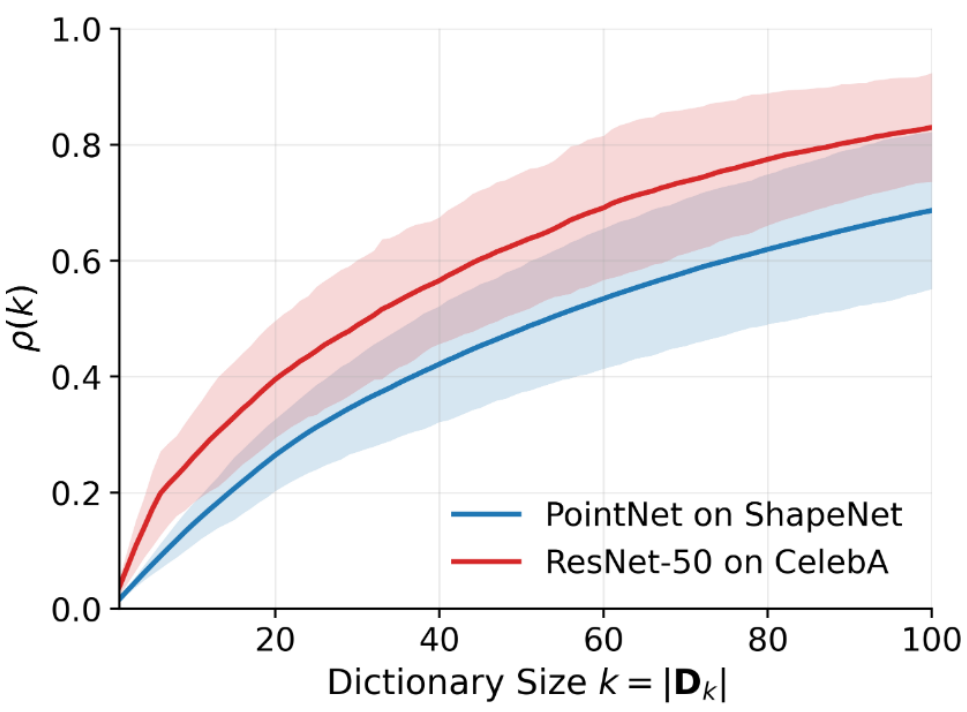

*Fig. 5* | ***Cross-sample transferability of salient interactions.*** *The curves show the explanation ratio $\rho(k)$ as a function of dictionary size $k$ for ResNet-50 on CelebA and PointNet on ShapeNet. For each model, the dictionary contains the $k$ most frequently occurring salient interactions across samples from the same category. The explanation ratio is the average proportion of salient interactions in an individual sample that are covered by the dictionary. Shaded areas indicate variation across samples. A dictionary of 100 interactions already explains 83.0% of salient interactions on CelebA and 68.7% on ShapeNet.*

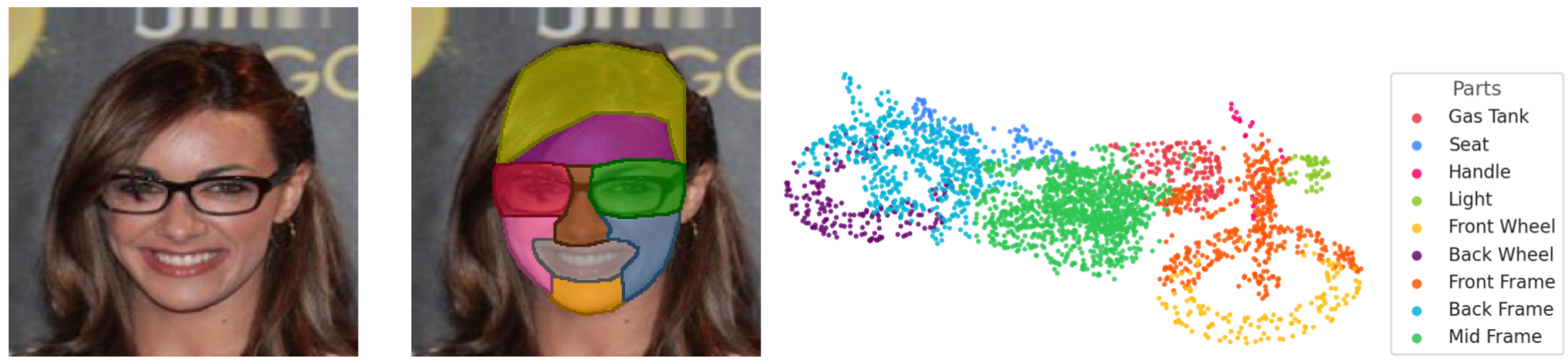

***Fig. 6* | *Examples of semantic-part annotations used to compute cross-sample interactions.*** *For CelebA, a face image is partitioned into nine annotated facial components, including the forehead, left and right eyes, nose, left and right cheeks, mouth, chin, and hair. For ShapeNet, a motorbike point cloud is partitioned into nine semantic parts: gas tank, seat, handle, light, front wheel, back wheel, front frame, mid frame, and back frame. Each annotated component or part is treated as one input variable when computing interactions.*

## Using interactions to explain the performance of an ANN

**Generalizability of interactions across different orders.** Because the output of an ANN can be decomposed into a sum of interaction effects, the overall generalizability of the ANN can be understood as the combined generalizability of its individual interactions. To assess the generalizability of interactions extracted from an ANN (namely, a *target model*), we follow the study[22] to learn a *baseline model*, which has the same architecture as the target model but is trained on the testing set. Given an input sample, a salient interaction extracted from the target model is considered *generalizable* if it is also identified as a salient interaction by the baseline model. Since the baseline model is trained directly on testing data, all interactions in the baseline model inherently reflect inference patterns that naturally generalize to the testing data.

Fig. 7(left) illustrates the proportion of generalizable interaction effect for each order. We observe that this proportion is usually high for low-order interactions (e.g., exceeding 70% for 1st-order interactions), but drops substantially for higher-order interactions (e.g., 32.75% for the 6th order). Moreover, 42.7% of positive interaction effect is generalizable, whereas this ratio decreases to 31.6% for negative interaction effect. This demonstrates that low-order and positive interactions are more generalizable, in contrast to their high-order and negative counterparts.

Given that low-order interactions are more generalizable, we ask whether these generalizable interactions emerge early in training. To this end, we analyze the interaction effect after 8 epochs. As shown in Fig. 7(right), the interaction effect is mainly concentrated in low-order terms, especially the first- and second-order interactions, which already exhibit strong generalizability. In contrast, higher-order interactions remain much weaker. This indicates that the model tends to learn simple and transferable interaction patterns at the early stage of training.

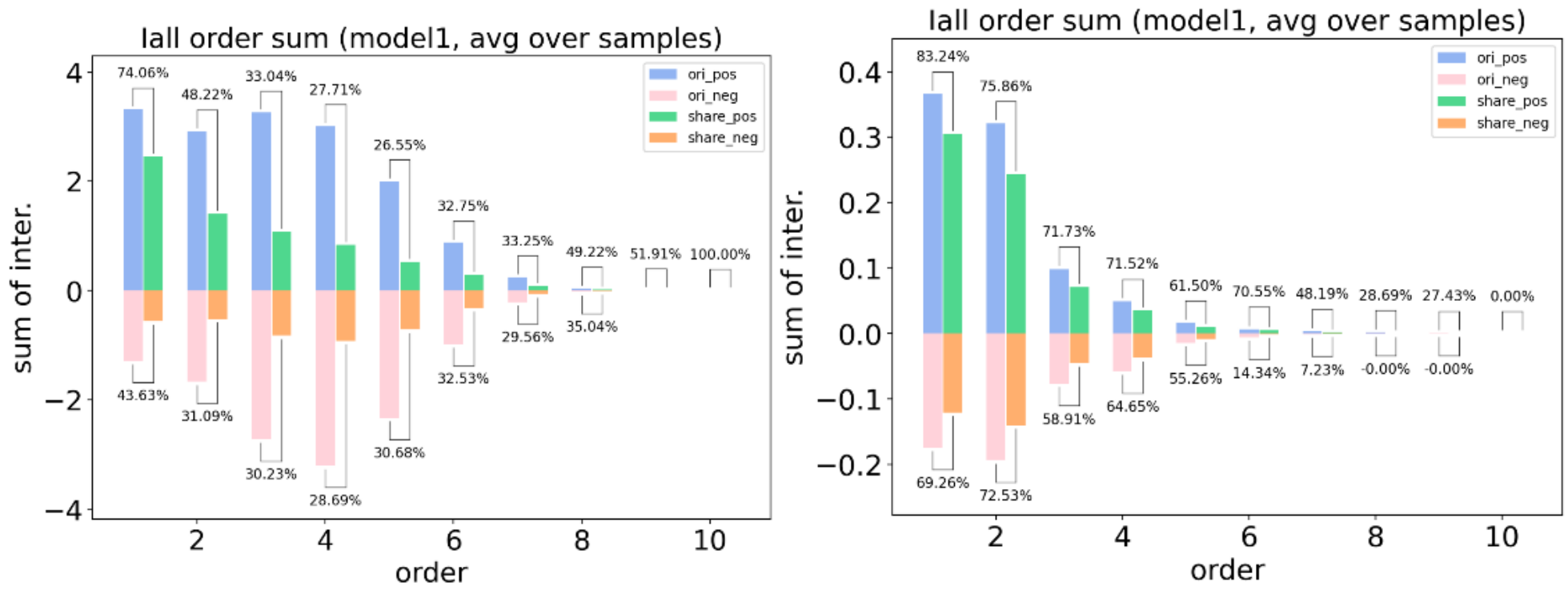

***Fig. 7* | *Order-wise generalizability of interactions at different training stages.*** *The left panel shows interaction effects extracted from models trained for 256 epochs, whereas the right panel shows those extracted after 8 epochs. At each order, positive and negative effects are decomposed into generalizable effects shared with the baseline model and non-generalizable effects specific to the target model. The annotated percentages indicate the proportions of generalizable positive and negative interaction effects. Low-order interactions are substantially more generalizable than high-order interactions and dominate at the early stage of training.*

**ANNs primarily learn medium- and high-order interactions during overfitting.** By tracking interaction strengths over time, we observe a two-phase dynamics (Fig. 8). Before the overfitting phase, the ANN predominantly reduces the strength of medium- and high-order interactions while amplifying low-order ones. Conversely, during the overfitting phase, the strength of medium- and high-order interactions steadily increases. These findings indicate a strong association between the learning of medium- and high-order interactions and the overfitting behavior of neural networks.

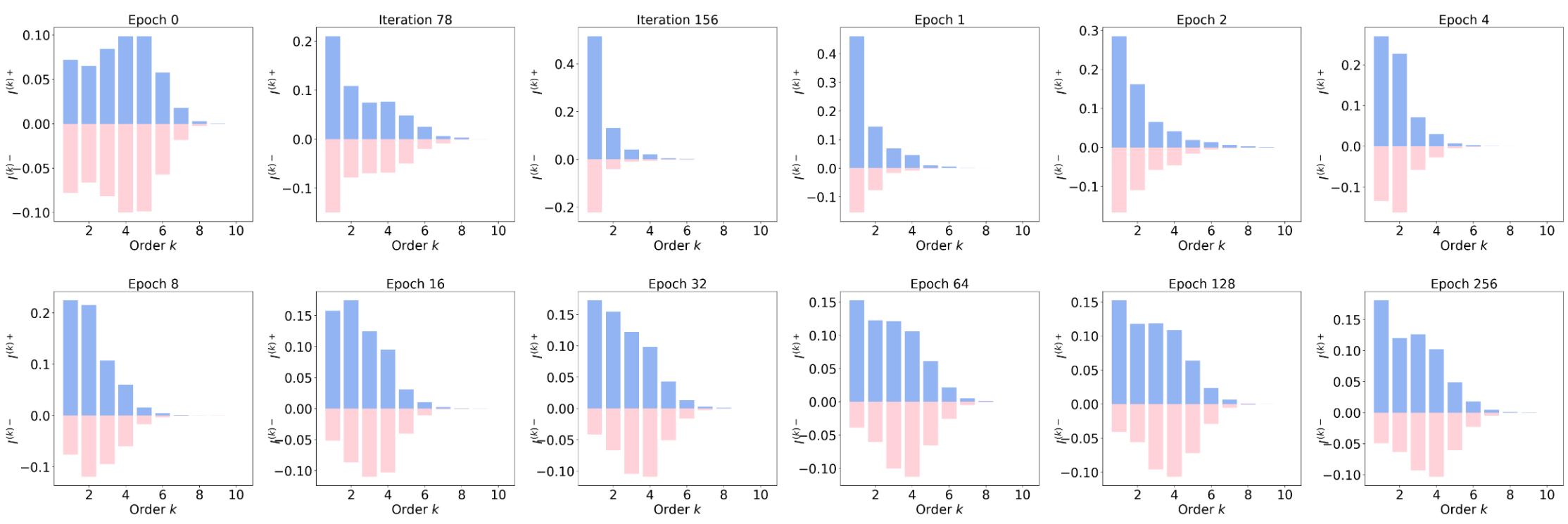

***Fig. 8* | *Two-phase dynamics of interaction learning during training.*** *The panels show the order-wise positive and negative interaction effects at different training epochs. During the early phase, the model progressively suppresses medium- and high-order interactions while strengthening low-order interactions. Then, medium- and high-order interaction effects increase again. These dynamics associate the late-stage learning of more complex interactions with the emergence of overfitting.*

## Discussion

The two mathematical criteria we have introduced suggest that the sparsity of interactions is not an artifact of a specific network architecture, but rather a natural consequence of the task's inherent representational demands. This is because the criteria are architecture-agnostic; they only require the ANN to be well-optimized for its task, making them likely to hold in practice. Furthermore, the observed cross-model and cross-sample transferability of interactions implies that interactions may serve as a fundamental explanation for ANNs.

Furthermore, we propose that these AND-OR interactions should be viewed as a powerful analytical lens to interpret the inference patterns of an ANN. For example, for ANNs towards regression tasks, the logical model projects the regression function to the space of interactions. In natural language processing, these interactions can be roughly interpreted as phrases automatically learned by the ANN, although these phrases may differ from human intuition.

In fact, AND-OR interactions are not the only possible representation of the inference patterns modeled by neural networks, and the ANN may encode various types of inference patterns. Our work simply projects the oracle inference patterns encoded by an ANN onto the space of interactions. Although the interaction-based explanation is not fully equivalent to the true function the ANN models, it provides a concise yet intuitive perspective to explain ANNs.

# Methods

## AND-OR Logical Model

**Definition.** Given an input sample $x = [x_1, \ldots, x_n]^\top$ with $n$ input variables indexed by $N = \{1,2,\ldots,n\}$, the explanation target is a scalar output $v(x) \in \mathbb{R}$ of the ANN, which can be manually specified. Typically, in multi-category classification tasks, $v(x)$ can be either the pre-softmax logit for the target class or the log odds of the target class. The logical model $h$ based on AND-OR interactions to explain inference logic on the given input $x$ is formulated as follows:

$$x' \in \Psi, \quad h(x') \coloneqq \sum_{S \in \Omega_{\text{and}}} I_S^{\text{and}} \cdot \mathbb{1}_{\text{and}}(S|x') + \sum_{S \in \Omega_{\text{or}}} I_S^{\text{or}} \cdot \mathbb{1}_{\text{or}}(S|x').$$

Here, $\Psi = \{x_S | S \subseteq N\}$ denotes a set of masked states generated based on the original input $x$, where $x_S$ means that all input variables in the set $N \backslash S$ are masked, while those in $S$ remain unmasked. For instance, given an input prompt to a Large Language Model (LLM), we mask an input token by replacing its embedding with a specific baseline embedding. Given an input image for image classification, we treat an image patch as an input variable to the ANN. As an implementation of masking an image patch, we replace feature vectors in all locations within the patch in an early-layer feature map, across all channels, with a baseline vector. The binary trigger function $\mathbb{1}_{\text{and}}(S|x')$ represents an AND interaction between input variables in the set $S$. This function returns 1 if and only if all variables in $S$ remain unmasked in $x'$. Similarly, $\mathbb{1}_{\text{or}}(S|x')$ represents an OR interaction, and it returns 1 if at least one variable in $S$ is unmasked in $x'$. $I_S^{\text{and}}$ and $I_S^{\text{or}}$ are effects for the AND interaction and the OR interaction, respectively. $\Omega_{\text{and}} \subseteq 2^N$ and $\Omega_{\text{or}} \subseteq 2^N$ denote the set of all AND interactions and the set of all OR-interactions, respectively.

Interaction effects in the AND-OR logical model are learned as follows, in which the LASSO-like objective function learns sparse interactions, and the predefined tolerance $\epsilon$ ensures the universal matching property on network outputs $v(x')$ on all masked states in $\Psi$.

$$\min_{\{I_S^{\text{and}}\},\{I_S^{\text{or}}\}} \sum_{S\subseteq N} |I_S^{\text{and}}| + |I_S^{\text{or}}| \quad \text{subject to} \quad \forall x' \in \Psi, |h(x') - v(x')| \leq \epsilon$$

In our experiments, we set $\epsilon = 0.02 \cdot \mathbb{E}_x[|v(x) - v(x_\emptyset)|]$.

**Order of interactions.** The order of an interaction quantifies the interaction's complexity. It is defined as the number of input variables in the set $S$, *i.e.*, $order(S) \triangleq |S|$. A low-order interaction represents a simple non-linear relationship among a few input variables, while a high-order interaction represents a simple non-linear relationship among manyinput variables

## Discovery of interaction sparsity

Although the above objective function is designed to encourage sparsity, the critical question remains whether the learned interactions are truly sparse in practice. We find that although there exists $2^n - 1$ combinations of the $n$ input variables in total (corresponding to $\emptyset \neq S \subseteq N$), only very few combinations $S$ have significant interaction effects $I_S^{\text{and}}$ or $I_S^{\text{or}}$.

**Models and datasets**. For image classification, we use VGG-16 and ResNet-18/34/50 models trained on the ImageNet dataset. We randomly sample one image from each category in the ImageNet validation set and obtain 1000 images to compute interactions.

For text generation, we use Large Language Models, including Qwen2.5-0.5B/1.5B/3B/7B/14B, Deepseek-R1-distilled-1.5B/7B/8B/14B, and Phi-4. We compute interactions on a dataset containing 100 sentences spanning various topics, which cover topics such as corporate restructuring, science and technology, economic policy, global trade, geopolitics, public health, political developments, cryptocurrency collapse, monetary policy, and climate and energy equity.

For 3D point cloud classification, we use PointNet trained on the ShapeNet dataset. We annotate semantic parts for 100 samples in the motorbike category. Our annotations are based on the original part annotations from the ShapeNet dataset, which include gas tank, seat, handle, light, wheel, and frame. Following prior work[23], we further refine them into a more fine-grained set of nine parts: gas tank, seat, handle, light, front wheel, back wheel, front frame, mid frame, and back frame.

**Details on computing interactions.** For image data, we consider image patches as input variables for the ANN. We follow previous study[24] to mask features in an low layer to generate a masked state $x' = x_S$. Specifically, we uniformly partition the feature map into 5×5 feature blocks. To ensure divisibility by 5, we discard 3 rows/columns on four edges of the feature map for ResNet models, and 1 row/column for VGG models. We then select 10 feature blocks with the largest average activation as the 10 input variables to calculate interactions. In this way, when we mask an input variable (a feature block), we replace all feature values within this block with a fixed baseline vector $b^* \in \mathbb{R}^c$, where $c$ is the number of channels. All values in the $i$-th channel are replaced by $b_i^*$. We use the feature map after the fourth ReLU layer for VGG models (with a spatial dimension of 112× 112), and the feature map after the final layer of the conv2_x stage for ResNet models (with a spatial dimension of 56×56). Feature blocks that are not in the set $N$ remain unmasked on all masked states $x_S$. We set $v(x_S)$ as the log odds of the ground truth category: $v(x_S) \triangleq \log \frac{p(y^{\text{truth}}|x_S)}{1-p(y^{\text{truth}}|x_S)}$.

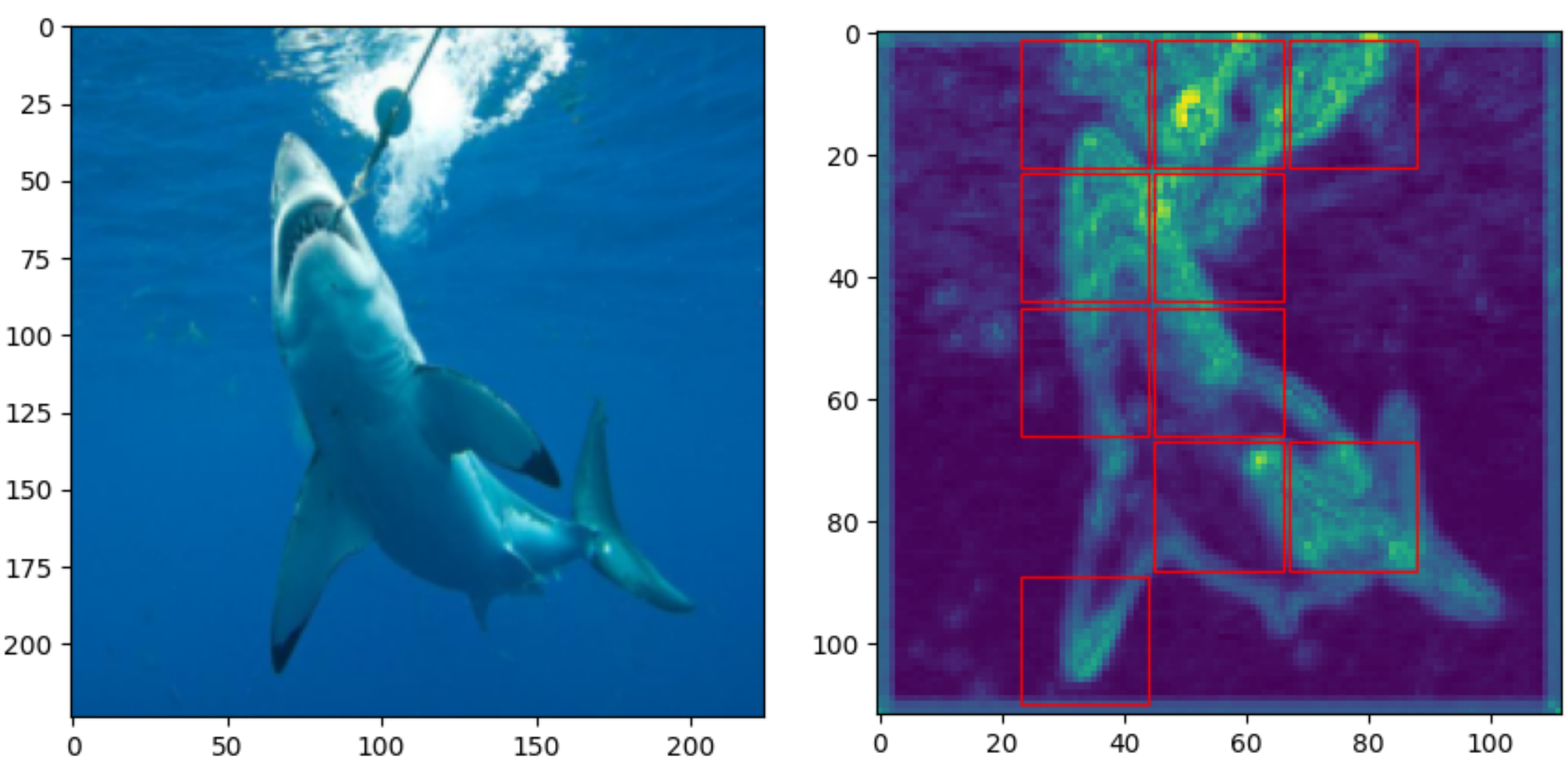

*Fig. 9 Example of feature block selection.*

For textual data, we treat each word or phrase in the input prompt as an input variable. Specifically, we manually select semantically meaningful phrases from the input prompt, where each phrase may consist of one or more tokens. Let us assume that the input prompt contains $t$ input variables in total. To generate the masked state $x' = x_S^{\leq t}$, we mask each phrase in $N\setminus S$ by replacing the embeddings of all the tokens in the phrase with a baseline embedding, which represents a no-information state. We set $v(x_S^{\leq t})$ as the log odds of next token predicted by the LLM: $v(x_S^{\leq t}) \triangleq \log \frac{p(x^{t+1}|x_S^{\leq t})}{1-p(x^{t+1}|x_S^{\leq t})}$.

For 3D point cloud objects in the motorbike category in the ShapeNet dataset, we treat each annotated semantic part as an input variable. To generate the masked state $x' = x_S$, we mask all semantic parts in $N\setminus S$ by moving these 3D points the center of the entire point cloud. We set $v(x_S)$ as the log odds of the ground truth category: $v(x_S) \triangleq \log \frac{p(y^{\text{truth}}|x_S)}{1-p(y^{\text{truth}}|x_S)}$.

**Evaluating the fidelity of the universal matching property with a few top-ranked interactions.** In this experiment, we simplify the logical model $h(x)$ by setting $\Omega_{\text{and}}$ and $\Omega_{\text{or}}$ to contain only 100 interactions with top-ranked absolute effects. For a set of input samples $\mathcal{X}$, the fidelity of the universal matching property is evaluated by the coefficient of determination $R^2 = 1 -$ $\frac{\sum_{x\in X}\sum_{x'\in\Psi(x)}\left(v(x')-\hat{h}(x')\right)^2}{\sum_{x\in X}\sum_{x'\in\Psi(x)}\left(v(x')-\bar{v}(x)\right)^2}$, where $\bar{v}(x)=\mathbb{E}_{x'\in\Psi(x)}[v(x')]$. Notably, we add a bias term $r$ to the simplified logical model $\hat{h}(x')=h(x')+r$ subject to $r=\mathbb{E}_{x'\in\Psi(x)}[v(x')-h(x')]$, which mitigates the effect of removing non-salient interactions.

## Mathematical criteria for the emergence of sparse interactions

For each input sample $x$, the two criteria for the sparsity of interactions are given as follows.

1. **Criterion 1 (Monotonicity):** Let us define the average $k$-order output $\bar{v}^{(k)}(x)\triangleq\mathbb{E}_{S\subseteq N:|S|=k}[v(x_S)]$ of the ANN as the average output over all masked states with $k$ unmasked input variables. The average $k$-order output $\bar{v}^{(k)}(x)$ must increase monotonically along with the increase of $k$, i.e., $\forall\, k<k', \bar{v}^{(k)}(x)<\bar{v}^{(k')}(x)$. Without loss of generality, we set $v(x_\emptyset)=0$ and $v(x_N)<1$ to constrain the scale of the average output.
2. **Criterion 2 (Smoothness):** The absolute value of the second-order finite difference of the average $k$-order output $\bar{v}^{(k)}(x)$ with respect to the number of unmasked variables $k$ is upper bounded by a constant $\frac{C}{n}$: $\left|\bar{v}^{(i+2)}(x)-2\bar{v}^{(i+1)}(x)+\bar{v}^{(i)}(x)\right|\le\frac{C}{n}, i=0,1,\dots,n-2$.

**Upper bound for the number of salient interactions.** Given a specific sequence of the average $k$-order output $\bar{v}^{(0)}(x),\bar{v}^{(1)}(x),\dots,\bar{v}^{(n)}(x)$, the upper bound of the number of salient interactions are given as $U=\frac{M}{\eta\epsilon}$. Here, $M$ is the optimal value of the optimization problem $\min\sum_{k=0}^{n}\left|A_{\text{and}}^{(k)}\right|+\left|A_{\text{or}}^{(k)}\right|$ s.t. $\forall 0\le i\le n, \left|\bar{v}^{(i)}(x)-\bar{h}^{(i)}(x)\right|\le\epsilon$, and $\eta=\frac{\sum_{k=0}^{n}\left|A_{\text{and}}^{(k)}\right|+\left|A_{\text{or}}^{(k)}\right|}{\sum_{S\subseteq N}\left|I_S^{\text{and}}\right|+\left|I_S^{\text{or}}\right|}$ measures the extent to which positive and negative interactions cancel each other, where $A_{\text{and}}^{(k)}=\sum_{S\subseteq N,|S|=k}I_S^{\text{and}}$ and $A_{\text{or}}^{(k)}=\sum_{S\subseteq N,|S|=k}I_S^{\text{or}}$. $\eta=0$ means that the interaction effects of each order fully cancel each other out. Experiments in the main text show that the value of $\eta$ is not extremely small in real ANNs. Please see Supplementary Note 1 for the detailed derivation of the upper bound and the computation of $\eta$.

Given a specific setting of $n$, $C$, and $\epsilon$, we uniformly sample 10,000 $\bar{v}^{(k)}(x)$ sequences that satisfy the monotonicity and smoothness criteria. Each sequence corresponds a group of ANNs. Please see Extended Data for the detailed sampling algorithm. Then, we compute the value of $M$ for each sampled sequence. We report the 95th percentile of $M$ across these sequences as an indicator of the relative magnitude of the upper bound on the interaction number.

## Detailed settings for validating the commonness of mathematical criteria for interaction sparsity

To empirically validate the prevalence of the two mathematical criteria for interaction sparsity (Monotonicity and Smoothness), we conduct analysis on the aforementioned CNNs, LLMs, and PointNet.

For CNNs, recall that we partition the feature map into a 5×5 grid, initially defining 25 feature blocks as potential input variables. However, calculating the average $k$-order outputs, $\bar{v}^{(k)}(x)$, requires evaluating the network output for all $2^n$ masked states, which is computationally intractable for n=25. To address this computational issue, we reduce the dimensionality of the input space by aggregating adjacent feature blocks. We pair neighboring blocks in the 5×5 grid, resulting in 12 input variables each composed of two original blocks, and one input variable consisting of the single remaining block. This procedure yields a manageable set of n=13 input variables for each image, so that we can exhaustively generate all $2^{13}$ masked states. For LLMs and PointNet, we simply use the original setting of input variables.

## Experimental details of testing convergence of interactions across ANNs

We use a pair of LLMs to predict the same target token for the same input prompt. The target token is selected as the one that is assigned top probability by both LLMs. Then, we follow prior work[25] to extract shared interactions from the two LLMs. If an AND interaction is extracted by both LLMs as a salient interaction and exhibits interaction effects with the same sign in both LLMs, then we consider it shared by two LLMs.

## Details of measuring cross-sample transferability of interactions

For face images in the CelebA dataset, we annotate facial components for 50 input images with the *eyeglasses* attribute. Specifically, for each image, we annotate nine facial components as nine input variables to the model: forehead, left and right eyes, nose, left and right cheeks, mouth, chin, and hair. For 3D point clouds in the ShapeNet dataset, we follow the annotations of semantic parts in the motorbike category as input variables, which are introduced in the **Details on computing interactions** paragraph.

We measure the cross-sample transferability of interactions as follows. Let $D_m$ denote a dictionary containing the top-$m$ most frequent salient interactions through different samples, where an interaction is considered salient if its absolute interaction effect ($|I_S^{\text{and}}|$ or $|I_S^{\text{or}}|$) exceeds $\epsilon$. Then, given a set of salient interactions $\Omega_x$ extracted from each input sample $x$, we measure the ratio of salient interactions that are covered by this dictionary as follows, in order to evaluate the transferability of interactions.

$$\rho(m) = \mathbb{E}_x\left[\frac{|D_m \cap \Omega_x|}{|\Omega_x|}\right]$$

$\rho(m)$ is averaged over different input samples. A larger value of $\rho(m)$ means that more salient interactions in each sample can be covered the dictionary, thereby suggesting better cross-sample transferability. Theoretically, when the dictionary is larger (having a larger $m$ value), it tends to cover a larger ratio of salient interactions.

## Details of quantifying generalizability of interactions of different orders

To assess the generalizability of interactions, we follow previous study[22] to train a pair of VGG-16

models, as follows. We first split the Tiny-ImageNet training set into two non-overlapping and equal-sized subsets. We regard the first subset as the training set, and the second subset as the testing set. The training set is used to train our *target model*, while the testing set is used to train the *baseline model*. Given an input sample, the interactions *shared* by both models, as elaborated in Section **Convergence of interactions across different ANNs**, are identified as *generalizable interactions*. Because all interactions in the baseline model are learned to represent testing data, the interactions that transfer from the standard model to the baseline model can be considered generalizable, i.e., representing a meaningful inference pattern for unseen testing data.

Let $S^{(k),+}$ and $S^{(k),-}$ denote the sum of all salient positive interactions and the sum of all salient negative interactions of the $k$-th order in the standard model, respectively. Accordingly, $S_{\text{shared}}^{(k),+}$ and $S_{\text{shared}}^{(k),-}$ represent the sum of all generalizable positive interactions and the sum of all generalizable negative interactions of the $k$-th order, respectively. Thus, the metric $r^{k,+} = \frac{S_{\text{shared}}^{(k),+}}{S^{(k),+}}$ quantifies the ratio of generalizable positive interactions to all positive interactions of the $k$-th order. The metric $r^{k,-} = \frac{S_{\text{shared}}^{(k),-}}{S^{(k),-}}$ is for negative interactions.

## Details of computing interaction dynamics

We train a VGG-16 model from scratch on the Tiny-ImageNet training set, and investigate the dynamics of interactions throughout the entire training process. The VGG-16 model is trained for 256 epochs using Stochastic Gradient Descent (SGD), with a learning rate of 0.01, a momentum of 0.9, a weight decay of 0.0001, and a batch size of 128. We randomly sample one image from each category in the Tiny-ImageNet validation set, resulting in a set of 200 images. Then, given each model checkpoint during the training process, we compute AND-OR interactions on these images.

For interactions of each k-th order, we compute the normalized strengths of the salient positive and negative interactions to distinguish their different effects. Specifically, we compute $I^{(k)+} = \mathbb{E}_x\left[\sum_{\text{type}\in\{\text{and,or}\}}\sum_{S:|S|=k, I_S^{\text{type}}\geq\epsilon} I_S^{\text{type}}\right]/Z$ , $I^{(k)-} = \mathbb{E}_x\left[\sum_{\text{type}\in\{\text{and,or}\}}\sum_{S:|S|=k, I_S^{\text{type}}\leq-\epsilon} I_S^{\text{type}}\right]/Z$ , where $Z = \mathbb{E}_{1\leq k'\leq n}\mathbb{E}_x\left[\sum_{\text{type}\in\{\text{and,or}\}}\sum_{S:|S|=k', \left|I_S^{\text{type}}\right|\geq\epsilon} I_S^{\text{type}}\right]$ denotes the normalizing constant. The normalization removes the effect of the explosion of output values during the training process and enables us to only analyze the relative distribution of interaction strength.